\documentclass{article}

\usepackage{microtype}
\usepackage{graphicx}
\usepackage{subcaption}
\usepackage[table]{xcolor}
\usepackage{booktabs} 
\usepackage[dvipsnames]{xcolor}
\usepackage[dvipsnames*,svgnames]{xcolor}
\usepackage{hyperref}
\usepackage{multirow}
\usepackage{graphicx}  

\usepackage[preprint]{icml2026}

\usepackage{amsmath}
\usepackage{amssymb}
\usepackage{mathtools}
\usepackage{amsthm}

\usepackage[capitalize,noabbrev]{cleveref}

\theoremstyle{plain}
\newtheorem{theorem}{Theorem}[section]
\newtheorem{proposition}[theorem]{Proposition}

\theoremstyle{definition}

\theoremstyle{remark}

\usepackage[textsize=tiny]{todonotes}

\usepackage{amsmath,amsfonts,bm}

\def\eqref#1{equation~\ref{#1}}

\def\1{\bm{1}}

\def\vtheta{{\bm{\theta}}}

\def\vr{{\bm{r}}}

\def\vy{{\bm{y}}}

\DeclareMathAlphabet{\mathsfit}{\encodingdefault}{\sfdefault}{m}{sl}
\SetMathAlphabet{\mathsfit}{bold}{\encodingdefault}{\sfdefault}{bx}{n}

\icmltitlerunning{Efficient RLVR Training via Weighted Mutual Information Data Selection}

\begin{document}

\twocolumn[
  \icmltitle{Efficient RLVR Training via Weighted Mutual Information Data Selection}



  \icmlsetsymbol{equal}{*}

  \begin{icmlauthorlist}
  \icmlauthor{Xinyu Zhou}{hkust-gz}
  \icmlauthor{Boyu Zhu}{ucl}
  \icmlauthor{Haotian Zhang}{ks}
  \icmlauthor{Huiming Wang}{ks}
  \icmlauthor{Zhijiang Guo}{hkust-gz,hkust}
  \end{icmlauthorlist}

  \icmlaffiliation{hkust-gz}{HKUST(GZ)}
  \icmlaffiliation{ucl}{UCL}
  \icmlaffiliation{ks}{Kuaishou Technology}
  \icmlaffiliation{hkust}{HKUST}

  \icmlcorrespondingauthor{Zhijiang Guo}{zhijiangguo@hkust-gz.edu.cn}
  \icmlcorrespondingauthor{Huiming Wang}{huiming\_wang@mymail.sutd.edu.sg}

  \icmlkeywords{Machine Learning, ICML}

  \vskip 0.3in
]



\printAffiliationsAndNotice{}  

\begin{abstract}
Reinforcement learning (RL) plays a central role in improving the reasoning and alignment of large language models, yet its efficiency critically depends on how training data are selected. Existing online selection strategies predominantly rely on difficulty-based heuristics, favouring datapoints with intermediate success rates, implicitly equating difficulty with informativeness and neglecting epistemic uncertainty arising from limited evidence. We introduce \textsc{InSight}, an \textbf{IN}formation-guided data \textbf{S}ampl\textbf{I}n\textbf{G} met\textbf{H}od for RL \textbf{T}raining, grounded in a weighted mutual information objective. By modeling data outcomes with Bayesian latent success rates, we show that expected uncertainty reduction decomposes into complementary difficulty- and evidence-dependent components, revealing a fundamental limitation of difficulty-only selection. Leveraging this observation, \textsc{InSight} constructs a stable acquisition score based on the mean belief of datapoints' success rather than noisy sampled outcomes, and naturally extends to multi-rollout settings common in reinforcement learning with verifiable rewards (RLVR). Extensive experiments demonstrate that \textsc{InSight} consistently achieves state-of-the-art performance and improves training efficiency, including a \textbf{+1.41} average gain on Planning \& Mathmatics benchmarks, \textbf{+1.01} improvement on general reasoning, and up to \textbf{$\sim$2.2x} acceleration, with negligible additional computational overhead.
\end{abstract}


\section{Introduction}
Reinforcement learning (RL) has emerged as a central paradigm for aligning Large Language Models (LLMs) with human preferences and improving their reasoning abilities \citep{shao2024deepseekmathpushinglimitsmathematical, deepseekai2025deepseekr1incentivizingreasoningcapability, zeng2025simplerlzooinvestigatingtamingzero, deepscaler2025}. Despite its success, RL is sensitive to the choice of training data samples \citep{parashar2025curriculumreinforcementlearningeasy, qu2025promptdifficultyonlinepredicted, shen2025bots, shen2025skyworkr1v3technicalreport}, and is also expensive in computations and memory usage for policy evaluation and updates, since it requires massive rollouts, especially in GRPO-style training \citep{deepseekai2025deepseekr1incentivizingreasoningcapability}. Training on static and uniformly sampled data distributions is inherently inefficient: substantial computation is wasted on tasks that the model has already mastered or that remain intractable given its current capability \citep{yu2025dapoopensourcellmreinforcement, bae2025onlinedifficultyfilteringreasoning}. Beyond increased training cost, this mismatch degrades optimization stability by reducing the effective batch size. Consequently, a key challenge in RL is to adaptively select tasks of appropriate difficulty, ensuring that learning remains efficient as the model’s capabilities evolve.

\begin{figure*}[!ht]
    \centering
    \includegraphics[width=0.98\linewidth]{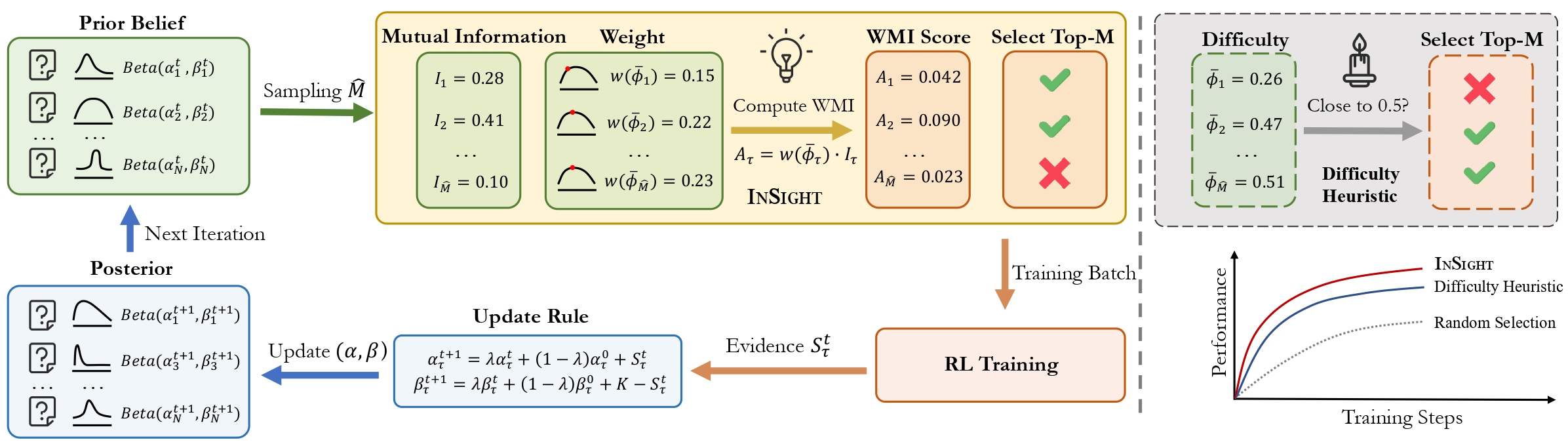}
    \caption{Overview of the \textsc{InSight} pipeline. \textsc{InSight} maintains a Bayesian belief over data success rates, scores candidate datapoints using Weighted Mutual Information (WMI), and selects the top-$M$ datapoints for RL training. Observed rewards update the posterior beliefs, enabling adaptive data selection that jointly accounts for data difficulty and accumulated evidence, unlike difficulty-only heuristics.}
    \label{fig:insight-pipeline}
    \vspace{-2mm}
\end{figure*}

A growing body of work has explored data selection strategies to improve the efficiency of RL training. Early approaches primarily adopt offline curricula \citep{parashar2025curriculumreinforcementlearningeasy, wen2025saristructuredaudioreasoning, shen2025skyworkr1v3technicalreport}, organizing training data along a predefined progression from easy to hard. While effective in controlled settings, such curricula are fixed prior to training and therefore do not adapt to the model’s evolving capabilities. Motivated by this limitation, several recent works propose online data selection methods \citep{qu2025promptdifficultyonlinepredicted, shen2025bots, yu2025dapoopensourcellmreinforcement} that dynamically select datapoints based on the current policy.

These methods, however, face an inherent trade-off between the cost of acquiring informative signals and the accuracy with which data utility can be estimated. Oversampling-based approaches \citep{yu2025dapoopensourcellmreinforcement, bae2025onlinedifficultyfilteringreasoning} address uncertainty by executing enlarged rollout batches to obtain reliable performance estimates, but this strategy introduces substantial computational overhead. In contrast, non-oversampling methods typically rely on a single proxy signal. In particular, difficulty-based heuristics \citep{shen2025bots, qu2025promptdifficultyonlinepredicted} prioritize datapoints whose empirical success rates lie near a target value (e.g., $0.5$), implicitly treating difficulty as a surrogate for informativeness. While such heuristics can be effective during early stages of training, their utility may diminish as evidence accumulates: data can remain difficult yet provide a limited additional learning signal once uncertainty has been sufficiently reduced. Consequently, existing approaches either incur significant evaluation costs to maintain accurate estimates or rely on simplified heuristics that do not fully reflect the evolving uncertainty structure during training.

To address the limitations, we propose \textsc{InSight}, an efficient information-aware data selection method available for all RL training, based on a weighted mutual information objective. Unlike prior online methods that rely on empirical or sampled success rates and implicitly equate difficulty with informativeness, \textsc{InSight} jointly considers both data difficulty and mutual information as shown in \autoref{fig:insight-pipeline}, which yields an acquisition score prioritizing datapoints that are expected to produce the greatest reduction in uncertainty. We further provide a theoretical analysis showing how information gain decays with accumulated evidence while preserving meaningful data-dependent structure. Across a range of benchmarks, \textsc{InSight} consistently achieves state-of-the-art performance (e.g., up to \textbf{+1.41} gain on Planning \& Mathmatics benchmarks and \textbf{+1.01} improvement on general reasoning), and improves the efficiency by \textbf{$\sim$2.2x} acceleration,  establishing a principled and effective alternative to difficulty-only data selection.



\vspace{-1mm}
\section{Related Work}
\noindent\textbf{RL for LLMs.}  RL has become a central paradigm for aligning LLMs with desired behaviors and unlock reasoning ability. Recently, Reinforcement Learning with Verifiable Rewards (RLVR) has demonstrated strong gains in structured reasoning domains~\citep{li2025system}, where reward signals can be automatically evaluated \citep{deepseekai2025deepseekr1incentivizingreasoningcapability, qwen2025qwen25technicalreport, deepcoder2025, yang2025qwen3technicalreport}. Among RL algorithms, GRPO \citep{shao2024deepseekmathpushinglimitsmathematical} has attracted increasing attention by eliminating the costly value network. Building on these foundations, recent work has focused on mitigating training instability and bias, reducing computational overhead, and improving sample efficiency \citep{yu2025dapoopensourcellmreinforcement, yue2025vapoefficientreliablereinforcement, yang2025depth, yang2025treerpo}. Furthermore, many recent studies have continued to advance the performance frontier across a wide range of domains and model scales \citep{luo2025codetestcasesenough, liang2025beyond, ma2025generalreasoner, meng2025mmeurekaexploringfrontiersmultimodal}, while complementary efforts focus on building scalable infrastructure for RL training \citep{Sheng_2025}. 

\noindent\textbf{Data Selection for RL.} Data selection has long been recognized as a critical factor in improving training efficiency and generalization in large-scale learning. A substantial body of work studies data selection in the context of pretraining and SFT, focusing on identifying informative, or high-quality samples \citep{xia2024lessselectinginfluentialdata, zhou2025hyperinfunleashinghyperpowerschulzs, kwon2024datainfefficientlyestimatingdata, koh2020understandingblackboxpredictionsinfluence, ghorbani2019datashapleyequitablevaluation}. RL is substantially more expensive than SFT due to repeated rollouts, policy evaluation, and credit assignment, making uniform sampling highly inefficient. As a result, selecting informative prompts is crucial for both training stability and computational efficiency. Early approaches adopt offline filtering, preselecting prompts based on heuristics such as difficulty or diversity \citep{li2025limrrlscaling, ye2025limoreasoning}. However, these methods lack adaptivity and often incur additional overhead for prompt assessment. To address this, recent work explores online data selection strategies that adapt to the current policy. Oversampling-based methods identify suitable data by executing oversized rollout batches and discarding uninformative samples \citep{yu2025dapoopensourcellmreinforcement,liu2025prorlprolongedreinforcementlearning}, but this substantially increases computational cost. To reduce evaluation overhead, several methods attempt to predict task success rates and select prompts accordingly. Some frame task selection as a non-stationary multi-armed bandit problem \citep{shen2025bots,qu2025promptdifficultyonlinepredicted}. However, these approaches typically rely on sampled or empirical success rates and implicitly equate difficulty with informativeness, neglecting how accumulated evidence reduces epistemic uncertainty over time. In contrast, our method \textsc{InSight} explicitly selects data by maximizing expected variance reduction. This principled formulation decouples difficulty from accumulated evidence, enabling adaptive and information-aware data selection without auxiliary rollouts or heuristic difficulty prediction.

\section{Preliminary}
\noindent\textbf{Setup \& Notations.} A datapoint $\tau$ in the reasoning tasks can be a form of a mathematical or logical problem, which belongs to the full pool of dataset $\mathcal{T}=\{\tau_i\}_{i=1}^N$. The parameters of the LLM at $t$-th training step is denoted as $\pi_{\vtheta_t}$. The selected data batch at $t$-th training step is $\mathcal{T}_t^M = \{\tau_{t,i}\}_{i=1}^M\subset \mathcal{T} $, where $M$ is the batch size. At $t$-th time step, for a datapoint $\tau$, the policy model $\pi_{\vtheta_t}$ generates $K$ independent response rollouts $\vy_{\tau}^t = \{y_{\tau}^{t,j}\}_{j=1}^K$.

\noindent\textbf{Data Difficulty Modeling.} For each datapoint $\tau$, we associate it with a success rate $\phi_{\tau}^t\in[0,1]$ to demonstrate its difficulty under the current policy, which is unknown and thus regarded as a latent variable \citep{shen2025bots,qu2025promptdifficultyonlinepredicted}. RLVR usually leverages a binary reward $r\in\{0,1\}$ as the training signal. Therefore, each response $y_{\tau}^{t,j}$ is examined with the ground-truth and scored with a binary reward:
\begin{equation}
    r_{\tau}^{t,j}\in \{0,1\}\sim \text{Bernoulli}(\phi_{\tau}^t)
\end{equation}
For each datapoint $\tau$, $\vr_{\tau}^t = \{r_{\tau}^{t,j}\}_{j=1}^K$ denotes the set of rewards for $K$ independent generated responses, and $\mathcal{R}_{t}^M = \{\vr_{\tau_{t,i}}^t\}_{i=1}^M$ denotes the collected rewards for the data batch at step $t$. Thus, the likelihood of observing $\vr_{\tau}^t$ given $\phi_{\tau}^t$ is binomial:
\begin{equation}
    p(\vr_{\tau}^t|\phi_{\tau}^t) = \binom{K}{S_{\tau}^t}(\phi_{\tau}^t)^{S_{\tau}^t}(1-\phi_{\tau}^t)^{K-S_{\tau}^t}
    \label{eqn:multi-rollouts-reward}
\end{equation}
where $S_{\tau}^t = \sum_{j=1}^Kr_{\tau}^{t,j}$ means the success counts. The entire optimization history up to step $t$ is noted as $\mathcal{H}_t = \{\mathcal{T}_{i}^M, \mathcal{R}_i^M\}_{i=0}^t$

\noindent\textbf{Bayesian Online Data Selection.} A Bayesian surrogate model is widely adopted to dynamically estimate the success rate $\phi_{\tau}^t$, and sample the most informative datapoints without requiring additional LLM inference \citep{shen2025bots, qu2025promptdifficultyonlinepredicted}. Each datapoint $\tau$ is modeled as an arm in a stochastic multi-armed bandit, with an unknown success rate $\phi_{\tau}^t\in[0,1]$. Selecting an arm corresponds to querying the current policy $\pi_{\vtheta_t}$ on datapoint $\tau$ and observing binary feedback $r_\tau \in \{0,1\}$ indicating success or failure. Unlike standard bandit formulations that aim to maximize cumulative reward, our objective is to adaptively select data that yield the most informative learning signals for updating the model  \citep{bae2025onlinedifficultyfilteringreasoning, chen2021evaluatinglargelanguagemodels}. 

A natural way to model the initial success rate is via a Beta prior distribution \citep{shen2025bots, qu2025promptdifficultyonlinepredicted}, due to the straightforward inference and closed-form posterior updates:
\begin{equation}
    \phi_{\tau}^0\sim \text{Beta}(\alpha_{\tau}^0,\beta_{\tau}^0)
\end{equation}
where $\alpha_{\tau}^0$ and $\beta_{\tau}^0$ represent the accumulated counts of successes and failures, respectively, which are typically set to $(1,1)$ for a uniform prior. Then, the posterior distribution of $\phi_{\tau}^t$ given observations is formulated as:
\begin{equation}
    p(\phi_{\tau}^t|\mathcal{H}_t)\propto p(\vr_{\tau}^t|\phi_{\tau}^t)\cdot p(\phi_{\tau}^t|\mathcal{H}_{t-1})
    \label{eqn:posterior-gamma}
\end{equation}
Note that $p(\phi_{\tau}^t|\mathcal{H}_{t-1})\sim \text{Beta}(\alpha_{\tau}^t,\beta_{\tau}^t)$ represents the conditional prior. The posterior of $\phi$ would also follow the Beta distribution, since the Beta distribution is conjugate to the Bernoulli likelihood in \autoref{eqn:multi-rollouts-reward}:
\begin{equation}
    \phi_{\tau}^t|\mathcal{H}_t \sim \text{Beta}(\alpha_{\tau}^t+S_{\tau}^t, \beta_{\tau}^t+K-S_{\tau}^t)
    \label{eqn:posterior-gamma-2}
\end{equation}
Then $\alpha_{\tau}^{t+1} = \alpha_{\tau}^t+S_{\tau}^t, \beta_{\tau}^{t+1} = \beta_{\tau}^t+K-S_{\tau}^t$ serve as the prior for the next step. The temporal discounting techniques are also applied for better stability:
\begin{equation}
\begin{aligned}
      \alpha_{\tau}^{t+1} &= \lambda\cdot \alpha_{\tau}^t + (1-\lambda)\cdot \alpha_{\tau}^0 +S_{\tau}^t,\\ \beta_{\tau}^{t+1} &= \lambda\cdot \beta_{\tau}^t + (1-\lambda)\cdot \beta_{\tau}^0 + K - S_{\tau}^t
      \label{eqn:alpha-beta-posterior-update}
\end{aligned}
\end{equation}
\vspace{-3mm}

Previous works \citep{bae2025onlinedifficultyfilteringreasoning, qu2025promptdifficultyonlinepredicted,shen2025bots} all claim that the problems with mid-level difficulty (i.e.,  near a target value $\phi^*\approx 0.5$) can yield the most informative gradients for RL training, from empirical experiments. As a result, those problems whose sampled success rate $\hat{\phi}_{\tau}^t$ is closest to $\phi^{*}$, will be ranked as top and selected for the following training. 

\section{Limitation of Difficulty-Only Data Selection}
\label{sec:limit-difficylty-data-selection}
While potentially effective in early training, difficulty-only heuristics primarily target \textbf{aleatoric uncertainty} (i.e., the intrinsic stochasticity of reward outcomes for a given datapoint) by favoring prompts with high outcome variability, implicitly equating difficulty with informativeness, but ignore the \textbf{epistemic uncertainty}, which arises from limited rollouts and captures uncertainty about the data latent success rate under the current policy \citep{chan2024estimatingepistemicaleatoricuncertainty, H_llermeier_2021}. As evidence accumulates, epistemic uncertainty about the underlying success rate diminishes, and prompts may remain difficult yet yield limited gains for parameter estimation. 


We posit that a more accurate estimation of $\phi_{\tau}^t$ leads to more data-efficient selection for policy optimization. In RLVR, policy gradients are driven by reward feedback from sampled trajectories. When the surrogate estimate of task success is noisy or poorly calibrated, the resulting learning signal becomes unreliable. Prioritizing datapoints that maximize expected uncertainty reduction in $\phi_{\tau}^t$ would promote the collection of reward signals that are more reliable and discriminative, yielding gradients that better reflect true task difficulty and thereby improving downstream policy learning performance. To quantify the improvement in the estimation of $\phi$ after observing the reward signal $r$, we consider the expected variance reduction for $\phi$ as follows:
\begin{equation}
    \Delta V(\tau) = \text{Var}_{\text{prior}}(\phi_{\tau}) - \mathbb{E}_r[\text{Var}_{\text{posterior}}(\phi_{\tau}|r)]
    \label{eqn:expected-variance-reduction-1}
\end{equation}
which captures the expected decrease in uncertainty of $\phi_{\tau}$ after observing all possible reward outcomes. Without loss of generality, we consider the binary reward signal and $K=1$, which is well-suited to most common RLVR settings and used to illustrate evidence-dependent decay. 

According to conditional expectation, we can expand \autoref{eqn:expected-variance-reduction-1} and simplify it as:
\begin{align}
     \Delta V(\tau) 
     &=\frac{\alpha\beta}{(\alpha+\beta)^2(\alpha+\beta+1)^2} = \frac{\textcolor{NavyBlue}{\bar{\phi}_{\tau}\cdot(1-\bar{\phi_{\tau}})}}{\textcolor{BrickRed}{(n+1)^2}}
      \label{eqn:expected-variance-reduction-2}
\end{align}
where $n=\alpha+\beta$ and $\bar{\phi}_{\tau}=\frac{\alpha}{\alpha+\beta}$ represents the mean of $\phi$'s distribution. The full derivation is included in \autoref{appdix:expected-variance-reduction}.

\begin{figure}[]
    \centering
    \includegraphics[width=0.9\linewidth]{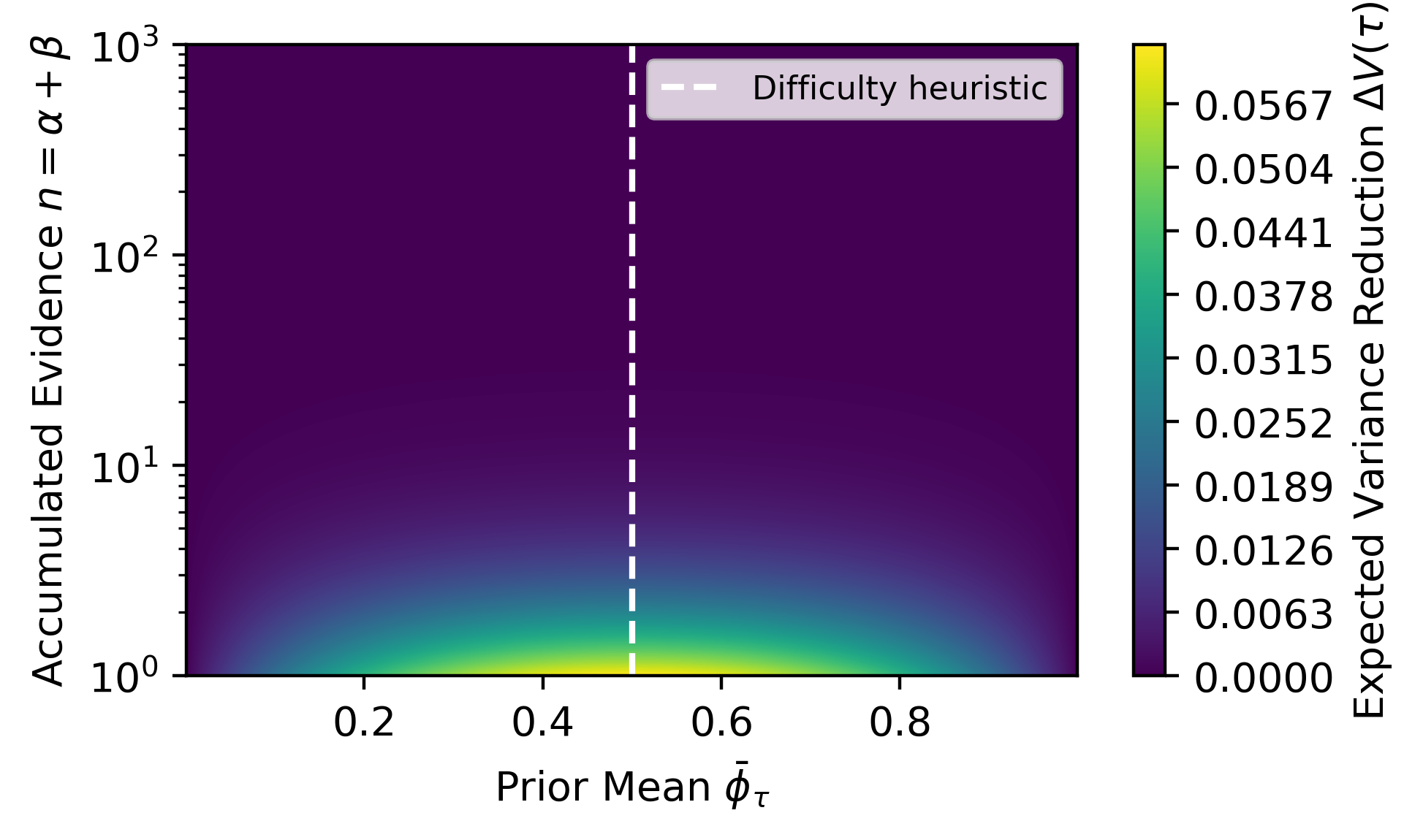}
    \caption{Expected variance reduction as a function of prior mean $\bar{\phi}_\tau$ and accumulated evidence $n$. While difficulty-based heuristics focus solely on $\bar{\phi}_\tau \approx 0.5$, the expected uncertainty reduction decays rapidly with evidence, revealing a fundamental limitation of difficulty-only selection.}
    \label{fig:exp-var-reduction-heatmap}
    \vspace{-3mm}
\end{figure}

\noindent\textbf{Analysis.} Existing difficulty-based data selection relies on a sampled success rate $\hat{\phi}_{\tau}^t$, and prioritize data whose sampled value lies closest to a target difficulty level $\phi^*$. However, as visualized in \autoref{fig:exp-var-reduction-heatmap}, expected variance reduction depends not only on $\bar{\phi}_{\tau}$ but also critically on the accumulated evidence $n=\alpha+\beta$. In particular, even when $\bar{\phi}_{\tau}\approx 0.5$, the contribution to uncertainty reduction becomes negligible once $n$ is large. In conclusion, while $\hat{\phi}_{\tau}^t$ concentrates around $\bar{\phi}_{\tau}$ in expectation, the difficulty heuristic ignores both sampling variability and the evidence-dependent decay, which would repeatedly select well-estimated yet mid-level difficult data, producing limited information gain in practice.

\vspace{-1mm}
\section{Method}
The preceding analysis exposes a key limitation of difficulty-only data selection: by relying on sampled success rates, purely difficulty-based heuristics conflate outcome variability with informativeness, optimizing only a partial aspect of uncertainty reduction. Our derivation shows that the distributional mean of the prior belief and accumulated evidence governs informativeness. Motivated by the decomposition of expected variance reduction into a difficulty-dependent and another evidence-dependent component, we decouple these two factors and design a new acquisition score:
\begin{equation}
    \mathcal{A}(\tau) \propto 
\underbrace{\textcolor{NavyBlue}{w(\cdot)}}_{\text{Aleatoric Exploitation}}\cdot  \underbrace{\textcolor{BrickRed}{I(\cdot)}}_{\text{Epistemic Exploration}} 
\label{eqn:acquisition-score}
\end{equation}
In the following, we detail the two components of the proposed acquisition score.

\vspace{-1mm}
\subsection{Mutual Information for Epistemic Exploration}
Mutual information provides a principled measure of the uncertainty reduction, as it directly captures the expected decrease in posterior uncertainty about $\phi_{\tau}$ induced by observations. Let $\Phi_{\tau}$ denote the random variable corresponding to the latent success rate $\phi_{\tau}$, and let $R$ denote the reward outcome. The mutual information between $R$ and $\Phi_{\tau}$ is defined as:
\begin{equation}
    I(R;\Phi_{\tau}) = H(\Phi_{\tau})-\mathbb{E}_{R}[H(\Phi_{\tau}|R=r)]
\end{equation}
This represents the expected reduction in uncertainty about $\Phi_{\tau}$ caused by observing the reward, indicating ``how informative will this datapoint be", which serves the Epistemic Exploration term in \autoref{eqn:acquisition-score}. $H(\cdot)$ denotes the entropy:
\begin{equation}
\begin{aligned}
     H(\Phi_{\tau})
     &=-\int_0^1f(\phi_{\tau})\ln{f(\phi_{\tau})}d\phi_{\tau}\\
     &=\ln{\text{B}(\alpha_{\tau},\beta_{\tau})}+(\alpha_{\tau}+\beta_{\tau}-2)\psi(\alpha_{\tau}+\beta_{\tau})\\ &\ \ \ \ \ -(\alpha_{\tau}-1)\psi(\alpha_{\tau})-(\beta_{\tau}-1)\psi(\beta_{\tau})
    \label{eqn:entropy}
\end{aligned}
\end{equation}
where $f(\cdot)$ denotes the probability density function, $\text{B}(\alpha,\beta)$ is the beta function, and $\psi(\cdot)$ denotes the digamma function. Full derivation is in  \autoref{appdix:entropy-beta-distribution}.

\noindent\textbf{Extend to multi-rollouts.} In practice, for example, using GRPO for RLVR training, each datapoint $\tau$ is typically evaluated multiple times to obtain $K$ independent responses. To account for this, we extend the above single-observation mutual information and variance reduction derivations to the multi-rollout setting.

Let $S_{\tau}=\sum_{k=1}^KR_k$ denote the number of successes observed, given $K$ independent responses with the corresponding rewards $R_{1:K}=(R_1,...,R_K)$ at a datapoint $\tau$, which follows the Binomial distribution as in \autoref{eqn:multi-rollouts-reward}. Then we further derive it as follows:
\begin{equation}
\scalebox{0.86}{$
\begin{aligned}
     P(S_{\tau}=s|K,\alpha_{\tau}, \beta_{\tau}) &= \int_0^1 P(S_{\tau}=s|\Phi_{\tau}=\phi,K) P(\Phi=\phi)d\phi\\
     &=\binom{K}{s}\frac{\text{B}(\alpha_{\tau}+s,\beta_{\tau}+K-s)}{\text{B}(\alpha_{\tau},\beta_{\tau})}
     \label{eqn:reward-binomial}
\end{aligned}
$}
\end{equation}
Therefore, the mutual information can be written as:
\begin{equation}
\scalebox{0.9}{$
\begin{aligned}
  &\textcolor{BrickRed}{I(R_{1:K}; \Phi_{\tau})}
   = H(\Phi_{\tau})-\mathbb{E}_{S_{\tau}}[H(\Phi_{\tau}\mid S_{\tau})]\\
  &= H(\Phi_{\tau}) - \sum_{s=0}^K
     P(S_{\tau}=s\mid K,\alpha_{\tau},\beta_{\tau})
     H(\Phi_{\tau}\mid S_{\tau}=s)\\
  &= H(\Phi_{\tau})-\sum_{s=0}^K
     \binom{K}{s}
     \frac{\mathrm{B}(\alpha_{\tau}+s,\beta_{\tau}+K-s)}
          {\mathrm{B}(\alpha_{\tau},\beta_{\tau})}
     H(\Phi_{\tau}\mid S_{\tau}=s)
     \label{eqn:mutual-information-multi-rollouts-1}
\end{aligned}
$}
\end{equation}

This expression demonstrates that the information gain from $K$ rollouts is governed by the expected reduction in posterior uncertainty about the latent success rate, averaged over all possible success counts.

\begin{algorithm}[tb]
   \caption{\textsc{InSight} Online Data Selection}
   \label{alg:wmi-algorithm}
\begin{algorithmic}
   \STATE {\bfseries Input:} Datapoint pool $\mathcal{T}=\{\tau\}_{i=1}^N$; Prior Beta parameters $\alpha,\beta$; Hyperparameters of weighted function ($w(\bar{\phi}_{\tau})$)  $\eta,\mu$;  Larger candidate batch size $\hat{M}$; selected batch size $M$; Policy model $\pi_{\vtheta_0}$; Total training steps $T$.
   \STATE {\bfseries Output:} Finetuned model $\pi_{\vtheta_T}$
    
    $\forall \tau\in \mathcal{T}$, initialize Beta parameters $(\alpha_{\tau}^0,\beta_{\tau}^0)\leftarrow(\alpha,\beta)$;
    \FOR{$t=0$ {\bfseries to} $T-1$}
    \STATE Randomly sample a larger candidate set $\mathcal{T}_t^{\hat{M}}\subset \mathcal{T}$;
    \FOR{$\hat{\tau}_{t,j}\in\mathcal{T}_t^{\hat{M}}$}
    \STATE Compute WMI score $\mathcal{A}(\hat{\tau}_{t,j})$ via \autoref{eqn:acquisition-score-2};
    \ENDFOR
    \STATE \textcolor{gray}{ \# {Rank Datapoints by WMI Score}}
    \STATE Select top datapoints set $\mathcal{T}_t^M\subset \mathcal{T}_t^{\hat{M}}$ by \autoref{eqn:top-WMI-selection};
    \FOR{$\tau_{t,i}\in\mathcal{T}_t^{{M}}$}
    \STATE \textcolor{gray}{\# Generate Responses and Compute Rewards}
    \STATE Generate responses $\vy_{\tau_{t,i}}^t = \{y_{\tau_{t,i}}^{t,j}\}_{j=1}^K$ by $\pi_{\vtheta_t}$;
    \STATE Compute rewards $\vr_{\tau_{t,i}}^t = \{r_{\tau_{t,i}}^{t,j}\}_{j=1}^K$;
    \ENDFOR
    \STATE Update $\vtheta_t$ to $\vtheta_{t+1}$ with an RL algorithm;
    \STATE \textcolor{gray}{\# Posterior Update}
    \FOR{$\tau_{t,i}\in\mathcal{T}_t^{{M}}$}
    \STATE Update $(\alpha_{\tau}^{t+1},\beta_{\tau}^{t+1})$ via \autoref{eqn:alpha-beta-posterior-update};
    \ENDFOR
    \ENDFOR
\end{algorithmic}
\end{algorithm}

\vspace{-1mm}
\subsection{Asymptotic Analysis of Mutual Information}
We next analyze the asymptotic behavior of the mutual information to characterize how epistemic uncertainty about the latent success rate decays as evidence accumulates. 

\begin{proposition}
    Let $\Phi_{\tau}\sim \text{Beta}(\alpha_{\tau},\beta_{\tau})$ denote the latent success rate of a datapoint $\tau$, and $R\in\{0,1\}$ be a Bernoulli reward conditioned on $\Phi_{\tau}$. Define $n_{\tau}=\alpha_{\tau}+\beta_{\tau}$. Then, as $n_{\tau}\rightarrow\infty$, we have
    \begin{equation}
        I(R;\Phi_{\tau})\approx \mathcal{O}(\frac{1}{n_{\tau}})
    \end{equation}
\end{proposition}
A complete derivation is provided in \autoref{appdix:asymptotic-scaling-mutual-information}. 

\noindent\textbf{Interpretation.} This result demonstrates that the information gained from an additional observation decreases as prior evidence accumulates, reflecting the diminishing epistemic uncertainty of well-understood datapoints.

\subsection{Weighted Function for Aleatoric Exploitation}
While mutual information provides a principled measure of epistemic uncertainty reduction, it is agnostic to task difficulty and treats all outcome regimes symmetrically. To account for this, we introduce a weighted function $w(\cdot)$ that modulates the epistemic acquisition term with a controlled preference for informative difficulty regimes.

\noindent\textbf{Motivation.} The design of it is motivated by two complementary factors. First, effective data selection should favor prompts with sufficient outcome variability, as near-deterministic regimes provide a limited learning signal. Second, learning dynamics in reinforcement learning are known to benefit from curriculum-like progression \citep{parashar2025curriculumreinforcementlearningeasy,bae2025onlinedifficultyfilteringreasoning}, and trivial tasks may mislead the model to exploit shortcuts to bypass meaningful reasoning \citep{laidlaw2025correlatedproxiesnewdefinition, parashar2025curriculumreinforcementlearningeasy}.

\begin{figure}
    \centering
    \includegraphics[width=0.98\linewidth]{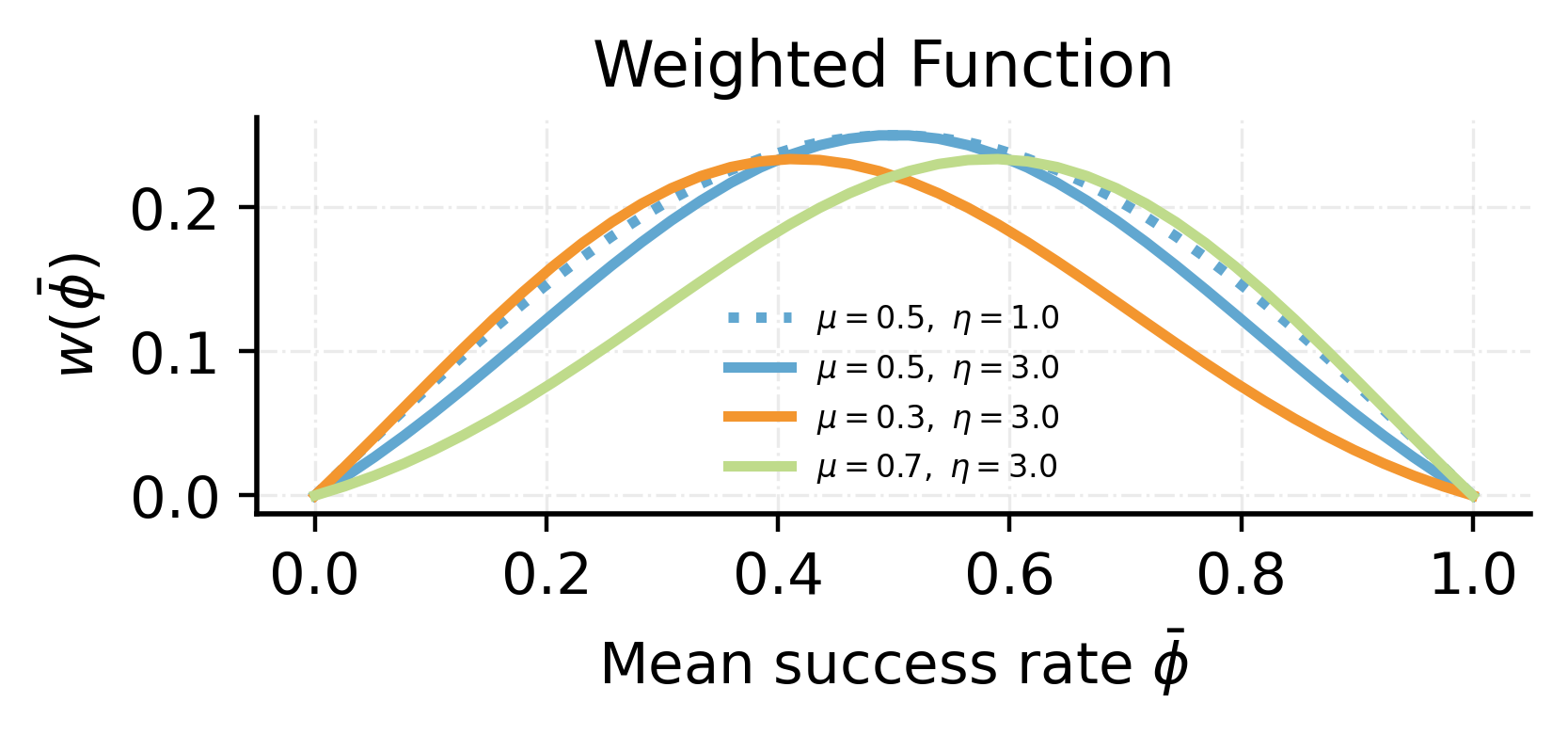}
    \caption{Weighted functions $w(\bar{\phi})$ under different $\eta, \mu$. }
    \label{fig:weighted-function}
    \vspace{-2mm}
\end{figure}

Inspired from \autoref{eqn:expected-variance-reduction-2}, unlike difficulty-only methods that operate on a sampled success rate, our weighting function is defined over the mean of the current prior belief of the latent success rate: $\bar{\phi}_{\tau}=\mathbb{E}[\phi_{\tau}]$, providing an expectation-based characterization of task difficulty. As a result, we design the weighted function as (shown in \autoref{fig:weighted-function}):
\begin{equation}
\textcolor{NavyBlue}{w(\bar{\phi}_{\tau})} =
\underbrace{\left(\bar{\phi}_{\tau}\left(1 - \bar{\phi}_{\tau}\right)\right)}_{\text{High Variance Filter}}
\cdot
\underbrace{\exp\!\left(-\eta\bigl(\bar{\phi}_{\tau}-\mu\bigr)^2\right)}_{\text{Bias to Difficulty }\mu}
\label{eqn:weighted-function}
\end{equation}
where the first term focuses on regions with high outcome variability, the second term introduces a smooth bias towards a beneficial difficulty level $\mu$, and another hyperparameter $\eta$ controls the sharpness of curriculum bias.

\begin{table*}[!ht]
\centering
\caption{Evaluation results across planning and mathematics benchmarks. Accuracy is computed as the average \texttt{pass@1} over 16 independent generations per problem. \textbf{Bold} indicates the best results, and \textcolor{gray}{gray} denotes the oracle baseline. {$\uparrow(\downarrow)$} indicates the improvement(degradation) compared to the \textsc{Random} baseline. } 
\label{tab:deepscaler-main-results}
\resizebox{\linewidth}{!}{%
\begin{tabular}{lllllllll}
\toprule
\textbf{\textsc{Models}}                      & \textbf{\textsc{Methods}}        & \textbf{\textsc{CountDown}}                   & \textbf{AIME24}              & \textbf{\textsc{AMC23}}               & \textbf{MATH500}             & \textbf{\textsc{Minerva.}}            & \textbf{\textsc{Olympiad.}}           & \textbf{\textsc{Avg.}}                \\ \midrule
                                     & \textsc{Random}           &71.57                       & 10.41                        & 43.90                        & 71.30                        & 21.60                        & 34.20                        & 42.16                        \\
    \textsc{Qwen3-0.6B}                                 & \textsc{MoPPS}          & 76.28                        & 10.83                        & 44.16                        & 70.60                        & 21.60                        & 34.25                        & 42.95                        \\
                                     & \textsc{Inverse-Evidence}    & 72.13                   & 10.60                        & 42.50                        & \textbf{71.80}               & 21.70                        & 34.80                        & 42.25                        \\
                                     
        & \textcolor{Tomato}{\textsc{InSight} (Ours)}                              &\textbf{76.70}\textsubscript{(5.13$\uparrow$)}  &\textbf{11.90}\textsubscript{(1.49$\uparrow$)}              & \textbf{44.30}\textsubscript{(0.40$\uparrow$)}               & 71.75\textsubscript{(0.45$\uparrow$)}                       & \textbf{21.94}\textsubscript{(0.34$\uparrow$)}               & \textbf{34.80}\textsubscript{(0.60$\uparrow$)}               & \textbf{43.56}\textsubscript{(1.40$\uparrow$)}              \\ \midrule
        
                                     & \textsc{Random}                              &  84.90  & 51.45                        & 78.01                        & 91.22                        & 41.13                        & 56.24                        & 67.16                        \\
           \textsc{ Qwen3-4B }                        & \textsc{MoPPS}       &      87.80                      & 50.00                        & 77.41                        & 90.82                        & 41.59                        & 56.45                        & 67.35                        \\
                                     & \textsc{Inverse-Evidence}   &83.59                     & 52.22                        & 77.18                        & \textbf{91.23}               & 41.49                        & 56.20                        & 66.99                        \\
                                     
          &  \textcolor{Tomato}{\textsc{InSight} (Ours)}                              &  \textbf{87.96}\textsubscript{(3.06$\uparrow$)}&\textbf{53.75}\textsubscript{(2.30$\uparrow$)}             & \textbf{79.00}\textsubscript{(0.99$\uparrow$)}        & 91.22\textsubscript{(0.00$\uparrow$)}                       & \textbf{41.80}\textsubscript{(0.67$\uparrow$)}               & \textbf{57.00}\textsubscript{(0.76$\uparrow$)}               & \textbf{68.46}\textsubscript{(1.30$\uparrow$)}             \\ \midrule
                                     & {\color[HTML]{9B9B9B} DS} &{\color[HTML]{9B9B9B} 83.05} &{\color[HTML]{9B9B9B} 50.62} & {\color[HTML]{9B9B9B} 80.79} & {\color[HTML]{9B9B9B} 90.82} & {\color[HTML]{9B9B9B} 38.39} & {\color[HTML]{9B9B9B} 53.38} & {\color[HTML]{9B9B9B} 66.17} \\
                                     & \textsc{Random} & 79.02                     & 46.25                        & 77.90                        & \textbf{90.50 }                       & 37.98                        & 51.80                        & 63.90                       \\
         \textsc{R1-Distill-Qwen-7B}                            & \textsc{MoPPS }   &          81.66                     & 47.20                        & 77.86                        & 90.40                        & 38.02                        & 52.01                        & 64.52                       \\
                                     & \textsc{Inverse-Evidence}   & 78.32                    & 47.00                        & 77.70                        & 90.40                        & 37.93                        & 52.07                        & 63.90                        \\
   &  \textsc{Expected-Difficulty} & 81.20	&47.50	&78.10	&90.43	&38.39	&52.25	&64.64 \\
                                  
 &  \textcolor{Tomato}{\textsc{InSight} (Ours)}                              &\textbf{81.73}\textsubscript{(2.71$\uparrow$)} &\textbf{47.71}\textsubscript{(1.46$\uparrow$)}                   & \textbf{78.40}\textsubscript{(0.50$\uparrow$)}                   & 90.45\textsubscript{(0.05$\downarrow$)}                         & \textbf{39.06}\textsubscript{(1.08$\uparrow$)}                         & \textbf{52.50}\textsubscript{(0.70$\uparrow$)}                        & \textbf{64.98}\textsubscript{(1.08$\uparrow$)}                        \\ \bottomrule
\end{tabular}
}
\end{table*}

\subsection{Weighted Mutual Information–Based Selection
}
Combining the preceding components, we define the final acquisition score, named Weighted Mutual Information (WMI), for a datapoint $\tau$ as:
\begin{equation}
    \mathcal{A}(\tau)=\textcolor{NavyBlue}{w(\bar{\phi}_{\tau})}\cdot \textcolor{BrickRed}{I(R_{1:K},\Phi_{\tau})}
    \label{eqn:acquisition-score-2}
\end{equation}
At each iteration, we will select the datapoints corresponding to the largest WMI score. Formally, given a larger candidate set $\mathcal{T}_t^{\hat{M}}$, we choose:
\begin{equation}
\mathcal{T}_t^{M}=\text{Top-}M\left(\left\{\tau\in\mathcal{T}_t^{\hat{M}}|\mathcal{A}(\tau)\right\}\right)
\label{eqn:top-WMI-selection}
\end{equation}
where $\hat{M}\gg M$. Then top-$M$ selected datapoints are used for rollouts and policy training. The full algorithm of \textsc{InSight} is described in Algorithm~\autoref{alg:wmi-algorithm}, which can be easily integrated with any RLVR algorithm.

\begin{table*}[!ht]
\centering
\caption{Evaluation results across general-reasoning benchmarks, via LM-Evaluation-Harness framework. \textbf{Bold} indicates the best results. {$\uparrow$} indicates the improvement compared to the \textsc{Random} baseline.}
\label{tab:general-reasoning}
\resizebox{\linewidth}{!}{%
    \begin{tabular}{llllllll}
\toprule
\multicolumn{1}{l}{\multirow{2}{*}{\textbf{\textsc{Models}}}}                     & \multicolumn{1}{l}{\multirow{2}{*}{\textbf{\textsc{Methods}}}}                     & \multicolumn{4}{c}{\textbf{\textsc{MMLU}}}             & \multicolumn{1}{l}{\multirow{2}{*}{\textbf{\textsc{GPQA}}}}          & \multicolumn{1}{l}{\multirow{2}{*}{\textbf{\textsc{Avg.}}}}                \\

\multicolumn{1}{l}{}     &     \multicolumn{1}{l}{}                         & \multicolumn{1}{l}{\textsc{Humanities}} & \multicolumn{1}{l}{\textsc{Other}} & \multicolumn{1}{c}{\textsc{Social-Sci.}} & \multicolumn{1}{l}{\textsc{STEM}}  &                              \\ \midrule

 & \textsc{Random }           &38.83	&47.22	&52.32	&40.95	&26.34	&41.13                         \\
    \textsc{Qwen3-0.6B}                                 & \textsc{MoPPS}  & 39.17	&47.12	&51.12	&38.69	&27.90	&40.80                                                      \\
                                     & \textsc{Inverse-Evidence}  & 38.55	&45.19	&49.76	&36.95	&{29.46}	&39.98                                       \\
                                     
        &  \textcolor{Tomato}{\textsc{InSight} (Ours)}           & \textbf{39.21}\textsubscript{(0.38$\uparrow$)}	&\textbf{47.34}\textsubscript{(0.12$\uparrow$)}	&\textbf{52.84}\textsubscript{(0.52$\uparrow$)}	&\textbf{41.80}\textsubscript{(0.85$\uparrow$)}	&\textbf{29.50}\textsubscript{(3.16$\uparrow$)}	&\textbf{42.14}\textsubscript{(1.01$\uparrow$)}             \\ \midrule

 & \textsc{Random }    &  48.63	&60.41	&64.09	&53.47	&30.20	&51.36                              \\
    \textsc{Qwen3-1.7B}                                 & \textsc{MoPPS}    & 48.63	&60.32	&64.38	&53.35	&30.10	&51.35                         \\
                                     & \textsc{Inverse-Evidence} & 48.80	&60.40	&64.41	&54.14	&29.69	&51.48                                        \\
                                     
        &  \textcolor{Tomato}{\textsc{InSight} (Ours)}                              & \textbf{49.00}\textsubscript{(0.37$\uparrow$)}	&\textbf{60.83}\textsubscript{(0.42$\uparrow$)}	&\textbf{65.00}\textsubscript{(0.91$\uparrow$)}	&\textbf{54.61}\textsubscript{(1.14$\uparrow$)}	&\textbf{30.36}\textsubscript{(0.16$\uparrow$)}	&\textbf{51.96}\textsubscript{(0.60$\uparrow$)}             \\ \midrule

 & \textsc{Random} &59.13	&71.07	&77.87	&68.25	&34.38	&62.14                                    \\
     & \textsc{MoPPS}                                   & 59.62 & 71.00	&78.00	&69.01	&33.26	&62.17                      \\
                               \textsc{Qwen3-4B}      & \textsc{Inverse-Evidence}  & 59.36	&71.12	&78.03	&68.79	&33.48	&62.16                                       \\
& \textsc{Expected-Difficulty} & 59.60	&\textbf{71.22}	&77.93	&68.63	&33.50	&62.17\\
        & \textcolor{Tomato}{\textsc{InSight} (Ours)}                              &  \textbf{59.70}\textsubscript{(0.57$\uparrow$)}	&{71.13}\textsubscript{(0.06$\uparrow$)}	&\textbf{78.10}\textsubscript{(0.23$\uparrow$)}	&\textbf{69.14}\textsubscript{(0.89$\uparrow$)}	&\textbf{35.04}\textsubscript{(0.66$\uparrow$)}	&\textbf{62.62}\textsubscript{(0.48$\uparrow$)}            \\ \midrule
    \end{tabular}
}
\vspace{-2mm}
\end{table*}



\vspace{-1mm}
\section{Experiments}
In this section, we conduct comprehensive experiments to evaluate various data selection approaches on RL training and validate the effectiveness of our proposed method.

\begin{figure*}[!ht]
    \centering
    \includegraphics[width=0.96\linewidth]{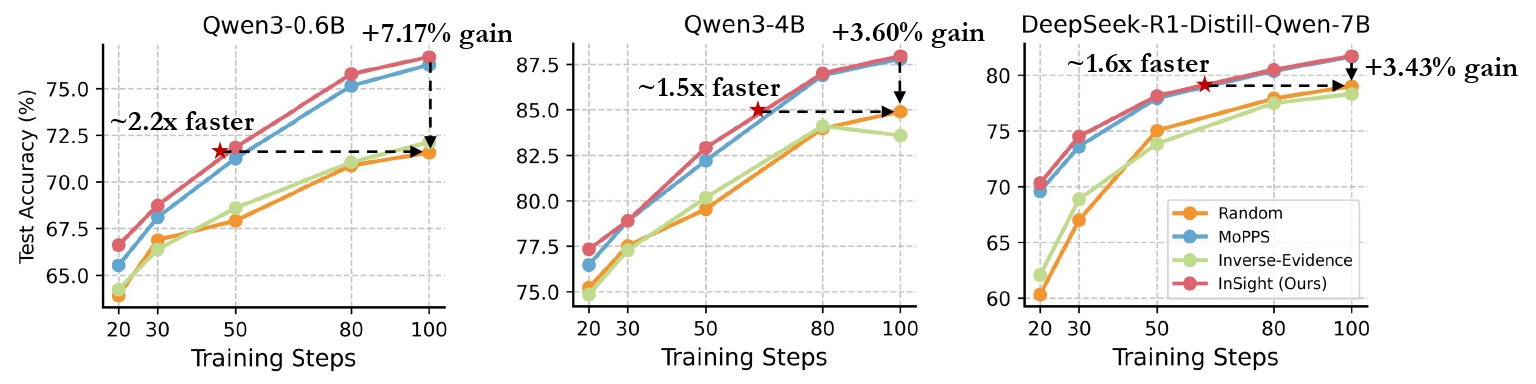}
    \caption{Performance comparisons among different methods on the Countdown task. Our proposed \textsc{InSight} outperforms the existing SOTA (\textsc{MoPPS}) and other baselines in both training efficiency and performance.}
    \label{fig:countdown-step}
    \vspace{-2mm}
\end{figure*}

\vspace{-1mm}
\subsection{Experimental Setup}
We evaluate \textsc{InSight} across representative reasoning tasks: planning, mathematics, and general reasoning. To examine its effectiveness, we adopt diverse LLMs with different scales, from 0.6B to 7B in size. For RL tuning, we choose one of the most commonly used algorithms: GRPO \citep{shao2024deepseekmathpushinglimitsmathematical}. Test accuracy is reported as average \texttt{pass@1} over 16 independent generations per problem. More implementation details are included in \autoref{appdix:implement-details}.

\vspace{-1mm}
\noindent\textbf{Reasoning Tasks.} (1) Planning: We adopt the Countdown task, which requires combining given numbers using basic arithmetic operations to reach a target value. Training is conducted on the subset of the Countdown dataset \citep{tinyzero}, and test is evaluated on the held-out split. (2) Mathematics: We train the LLMs on the DeepScaler training dataset \citep{deepscaler2025}, which consists of approximately 40.3K mathematics problem-answer pairs. Then we evaluate the final performance on common benchmarks, including AIME24, AMC23, MATH500 \citep{lightman2023lets}, Minerva Math (Minerva.) \citep{lewkowycz2022solvingquantitativereasoningproblems}, and OlympiadBench (Olympiad.) \citep{he2024olympiadbenchchallengingbenchmarkpromoting}. (3) General Reasoning: We utilize the WebInstruct-verified dataset \citep{ma2025generalreasoner}, which encompasses a broad range of domains (e.g., physics, chemistry, finance), and evaluate the performances on MMLU and GPQA-Main \citep{hendrycks2021ethics, rein2023gpqagraduatelevelgoogleproofqa}.

\noindent\textbf{Models.} For Countdown and Mathematics tasks, we include three LLMs in different scales: Qwen3-0.6B, Qwen3-4B \citep{yang2025qwen3technicalreport}, and DeepSeek-R1-Distil-Qwen-7B; and for General Reasoning, we also use three LLMs: Qwen3-0.6B, Qwen3-1.7B, and Qwen3-4B \citep{yang2025qwen3technicalreport}.

\noindent\textbf{Baselines.}
(1) \textbf{\textsc{Random}}, which selects datapoints uniformly at random, serving as a non-adaptive baseline; (2) \textbf{\textsc{MoPPS}} \citep{qu2025promptdifficultyonlinepredicted}, a recent state-of-the-art Bayesian online sampling method, prioritizes datapoints whose estimated success rates are closest to targeted difficulty (i.e., $0.5$); (3) \textbf{\textsc{Inverse-Evidence}}, which is a theory-inspired heuristic baseline derived from the expected variance reduction in \autoref{eqn:expected-variance-reduction-2}. It selects datapoints based solely on the inverse of accumulated evidence, favoring prompts with smaller $n_{\tau}$, under the intuition that less-observed datapoints carry higher epistemic uncertainty, while ignoring task difficulty; (4) \textbf{\textsc{Expected-Difficulty}}, which selects datapoints whose mean success rate is closest to the target difficulty $0.5$, removing sampling variance in \textsc{MoPPS} but still ignoring evidence accumulation; and (5) \textbf{Dynamic Sampling (DS)} \citep{yu2025dapoopensourcellmreinforcement}, which over-samples datapoints and filter out those without effective gradients based on their exact evaluation. It is worth noting that DS is extremely time-consuming, and we view it as an oracle baseline rather than outperforming it in test accuracy (refer to \autoref{appdix:performance-runtime-comparison} for the performance and runtime cost comparisons in detail).

\subsection{Main results}
\vspace{-1mm}

\noindent\textbf{Planning \& Mathematics.} \autoref{tab:deepscaler-main-results} reports results on planning and mathematical reasoning benchmarks for models from 0.6B to 7B parameters. Across all scales, \textsc{InSight} consistently outperforms \textsc{Random} and \textsc{MoPPS}, achieving the highest average accuracy in every setting. Gains are most pronounced on CountDown and AIME24, with improvements of up to \textbf{+5.13} and \textbf{+2.30} over \textsc{Random}, respectively. The largest average improvement occurs on the 0.6B model (\textbf{+1.40}), with gains gradually diminishing as model size increases, though \textsc{InSight} still yields a \textbf{+1.08} average improvement on R1-Distill-Qwen-7B.

\vspace{-1mm}
\noindent\textbf{General Reasoning.} \autoref{tab:general-reasoning} presents results on MMLU and GPQA. While absolute improvements are smaller than in mathematical reasoning, \textsc{InSight} consistently achieves the strongest or near-strongest performance across all model sizes. Notable gains are observed on MMLU-STEM and GPQA, reaching \textbf{+1.14} on Qwen3-1.7B and \textbf{+3.16} on Qwen3-0.6B, respectively.

\vspace{-1mm}
\noindent\textbf{Discussions.} First, improvements decrease with model scale across all benchmarks, reflecting reduced headroom as larger models exhibit stronger priors and reasoning capabilities. Second, \textsc{Expected-Difficulty} consistently outperforms \textsc{MoPPS}, indicating that using the mean difficulty provides a more stable acquisition signal than sampling-based heuristics. In contrast, \textsc{Inverse-Evidence} performs comparably or even worse than \textsc{Random}, suggesting that epistemic uncertainty alone is insufficient when decoupled from task difficulty. On general reasoning tasks, where outcome noise is higher, \textsc{Random} occasionally surpasses \textsc{MoPPS}, highlighting the sensitivity of sampling-based difficulty heuristics to noisy surrogate signals.

\vspace{-0.5mm}
Finally, we evaluate training efficiency on CountDown, an in-domain benchmark drawn from the same distribution as the RL training data, providing a controlled setting for comparing training efficiency. As shown in \autoref{fig:countdown-step}, \textsc{InSight} reaches competitive or superior performance with fewer training steps, yielding up to \textbf{$\sim$2.2x}, \textbf{$\sim$1.5x}, and \textbf{$\sim$1.6x speedups} on Qwen3-0.6B, Qwen3-4B, and R1-Distill-Qwen-7B, respectively. Although \textsc{MoPPS} exhibits similar convergence behavior, \textsc{InSight} demonstrates more consistent early-stage gains and slightly stronger final performance, indicating that jointly modeling difficulty and epistemic uncertainty improves gradient efficiency.

\vspace{-4mm}
\section{Ablation Analysis}

\subsection{Effects of WMI Components}
We first analyze the contribution of each component (e.g., weighted function \textcolor{NavyBlue}{$w(\bar{\phi}_{\tau})$} and mutual information \textcolor{BrickRed}{$I(R_{1:K},\Phi_{\tau})$}) in the WMI score of \textsc{InSight}.

\vspace{-1mm}
\noindent\textbf{Results.} \autoref{tab:wmi-component-analysis} reports the component-wise ablation. Mutual information alone consistently underperforms the full WMI objective. While MI captures epistemic uncertainty in the latent success rate, it does not account for whether this uncertainty lies in difficulty regimes that yield informative policy gradients. Conversely, the difficulty weighting alone performs competitively but lacks explicit uncertainty awareness. Combining both components yields the most consistent performance across all benchmarks, demonstrating that jointly modeling epistemic uncertainty and task difficulty is crucial for effective data selection in RL.

\begin{table}[]
\centering
\caption{Evaluation comparisons of WMI components across different models. \textbf{Bold} indicates the best results.} 
\label{tab:wmi-component-analysis}
\resizebox{\linewidth}{!}{%
\begin{tabular}{lcccc}
\toprule
\textbf{\textsc{Models}}                      & \textbf{\textsc{Component}}                       & \textbf{AIME24}                   & \textbf{\textsc{AMC23}}           & \textbf{\textsc{Olympiad.}}                \\ \midrule
            & \textcolor{BrickRed}{$I(R_{1:K},\Phi_{\tau})$} & 9.37 & 43.52 & 34.01\\
    \textsc{Qwen3-0.6B}    & $\textcolor{NavyBlue}{w(\bar{\phi}_{\tau})}\cdot \textcolor{BrickRed}{I(R_{1:K},\Phi_{\tau})}$ & \textbf{11.90} & \textbf{44.30} & \textbf{34.80} \\ \midrule
     & \textcolor{BrickRed}{$I(R_{1:K},\Phi_{\tau})$} & 51.66 & 76.43 & 56.24\\
    \textsc{Qwen3-4B}    & $\textcolor{NavyBlue}{w(\bar{\phi}_{\tau})}\cdot \textcolor{BrickRed}{I(R_{1:K},\Phi_{\tau})}$ & \textbf{53.75} & \textbf{79.00} & \textbf{57.00} \\ \midrule
     & $\textcolor{NavyBlue}{w(\bar{\phi}_{\tau})}$ & 46.67 & 77.33 & 52.25 \\
    \textsc{R1-Distill-7B}   & \textcolor{BrickRed}{$I(R_{1:K},\Phi_{\tau})$}   & 46.67 & 77.90 & 52.35 \\
    & $\textcolor{NavyBlue}{w(\bar{\phi}_{\tau})}\cdot \textcolor{BrickRed}{I(R_{1:K},\Phi_{\tau})}$ & \textbf{47.71} & \textbf{78.40} & \textbf{52.50}
                                                   \\ \bottomrule
\end{tabular}
}
\vspace{-2mm}
\end{table}

\vspace{-1mm}
\subsection{Expected vs. Sampled Difficulty in WMI}
According to \autoref{eqn:expected-variance-reduction-2} and \autoref{eqn:weighted-function}, WMI relies on the mean of the underlying success-rate distribution. In this experiment, we explore the effect of replacing $\bar{\phi}_\tau$ with $\hat{\phi}_{\tau}\sim\text{Beta}(\alpha_{\tau},\beta_{\tau})$, which is always used in previous work.

\vspace{-1mm}
\noindent\textbf{Results.} As shown in \autoref{tab:gamma-analysis}, using the mean success rate $\bar{\phi}_{\tau}$ consistently outperforms the sampled variant $\hat{\phi}_{\tau}$ across all models and benchmarks. This aligns with our theoretical analysis: expected variance reduction depends on the mean belief and accumulated evidence, whereas sampling introduces unnecessary noise that destabilizes data ranking. The results confirm that mean-based estimation yields a more reliable and effective acquisition signal than sampled success rates used in prior work.

\begin{table}[]
\centering
\caption{Evaluation comparisons between $\bar{\phi}_{\tau}$ and $\hat{\phi}_{\tau}$ variants in WMI across different models. \textbf{Bold} indicates the best results.} 
\label{tab:gamma-analysis}
\resizebox{\linewidth}{!}{%
\begin{tabular}{lcccc}
\toprule
\textbf{\textsc{Models}}                      & $\bar{\phi}_{\tau}$ / $\hat{\phi}_{\tau}$                        & \textbf{AIME24}                   & \textbf{\textsc{AMC23}}           & \textbf{\textsc{Olympiad.}}                \\ \midrule
            & $\hat{\phi}_{\tau}$   & 10.41 & 43.22 & 34.25 \\
    \textsc{Qwen3-0.6B}    & $\bar{\phi}_{\tau}$ & \textbf{11.90} & \textbf{44.30} & \textbf{34.80} \\ \midrule
     & $\hat{\phi}_{\tau}$ & 51.45 & 77.48 & 56.33 \\
    \textsc{Qwen3-4B}    & $\bar{\phi}_{\tau}$ & \textbf{53.75} & \textbf{79.00} & \textbf{57.00}   \\ \midrule
     & $\hat{\phi}_{\tau}$ & 46.25 & 77.93 & 52.37\\
    \textsc{R1-Distill-7B}   & $\bar{\phi}_{\tau}$ & \textbf{47.71} & \textbf{78.40} & \textbf{52.50} 
    
                                                   \\ \bottomrule
\end{tabular}
}
\vspace{-3mm}
\end{table}

\vspace{-1mm}
\subsection{Effect of Difficulty Bias $\mu$ in WMI}
The WMI objective incorporates a difficulty bias that favors tasks near a target success rate $\mu$. We study the sensitivity of WMI to this design choice by varying $\mu$, assessing both robustness and the trade-off between exploration and curriculum shaping.

\begin{table}[]
\centering
\caption{Evaluation comparisons of different $\mu$ in weighted function across different models. \textbf{Bold} indicates the best results.} 
\label{tab:mu-analysis}
\resizebox{0.95\linewidth}{!}{%
\begin{tabular}{lcccc}
\toprule
\textbf{\textsc{Models}}                      & $\mu$                        & \textbf{AIME24}                   & \textbf{\textsc{AMC23}}           & \textbf{\textsc{Olympiad.}}                \\ \midrule
            & $0.1$ & 10.41	& \textbf{44.80} &	34.75 \\
    \textsc{Qwen3-0.6B}    & $0.3$ & \textbf{11.90} & {44.30} & \textbf{34.80} \\
     & $0.7$ & 10.41	& 42.62	&	33.09 \\ \midrule
     & $0.1$ & 50.83&	77.03&	56.32\\
    \textsc{Qwen3-4B}    &  $0.3$ & \textbf{53.75} & \textbf{79.00} & \textbf{57.00}   \\
     &  $0.7$ & 52.91 & 77.41 & 56.58  \\ \midrule
     & $0.1$ & 46.87 & 78.23 & 52.08 \\
    \textsc{R1-Distill-7B}   & $0.3$ & {47.71} & {78.40} & \textbf{52.50}\\
    & $0.7$ & \textbf{47.91}& \textbf{78.46} & 52.25  \\ \bottomrule
\end{tabular}
}
\vspace{-2mm}
\end{table}

\vspace{-1mm}
\noindent\textbf{Results.} \autoref{tab:mu-analysis} shows LLMs' behavior under the choices of the difficulty bias $\mu$. Generally, $\mu=0.3$ achieves the strongest or near-strongest performance, indicating that prioritizing moderately challenging tasks provides an effective balance between exploration and curriculum shaping. In contrast, more extreme settings ($\mu=0.1$ or $0.7$) tend to underperform, suggesting that overemphasizing very easy or very difficult tasks is less effective, and this phenomenon is more significant in smaller models (e.g., Qwen3-4B's AIME24 score drops to 50.83 from 53.75 when $\mu=0.1$). The sensitivity to $\mu$ further diminishes as model size increases, with the 7B model exhibiting only minor variations, reflecting greater inherent robustness. We further analyze the impact of the candidate batch size $\hat{M}$ on model performance, with a detailed discussion provided in \autoref{appdix:hatM-analysis}.

\vspace{-1mm}
\section{Conclusion}
\vspace{-1mm}
We introduced \textsc{InSight}, an information-guided data selection framework for efficient reinforcement learning of large language models. By analyzing expected variance reduction, we revealed why difficulty-only heuristics fail to capture true informativeness as evidence accumulates, and motivated a weighted mutual information objective that jointly accounts for task difficulty and epistemic uncertainty. \textsc{InSight} operationalizes this insight using stable mean posterior beliefs and naturally extends to multi-rollout RLVR settings. Experiments across planning, mathematical, and general reasoning benchmarks show that \textsc{InSight} consistently improves training efficiency and final performance over strong online baselines, particularly in smaller-model and low-resource regimes. These results demonstrate that principled, information-aware data selection offers a robust alternative to heuristic difficulty-based sampling.
\section*{Impact Statements}
This paper presents work whose goal is to advance the field of machine learning by improving the efficiency of reinforcement learning through principled data selection. The proposed method focuses on algorithmic efficiency and does not introduce new model capabilities or application domains. While more efficient training may reduce computational costs and resource usage, we do not identify any specific societal or ethical risks beyond those commonly associated with large language model development.

\bibliography{example_paper}
\bibliographystyle{icml2026}

\newpage
\appendix
\onecolumn

\section{Derivation of Expected Variance Reduction }
\label{appdix:expected-variance-reduction}
In this section, we include the complete derivation of expected variance reduction in \autoref{sec:limit-difficylty-data-selection}. We know that $\phi\sim \text{Beta}(\alpha,\beta)$, and the variance of it is:
\begin{equation}
    \text{Var}(\phi) = \frac{\alpha\beta}{(\alpha+\beta)^2(\alpha+\beta+1)}
    \label{eqn:beta-variance}
\end{equation}
For the posterior distribution of $\phi$, we first apply the Bayes' rule:
\begin{equation}
    p(\phi|r)=\frac{p(r|\phi)p(\phi)}{p(r)} \propto p(r|\phi)p(\phi)
\end{equation}
The likelihood for a single reward observation $r\in\{0,1\}$ is:
\begin{equation}
    p(r|\phi) = \phi^r(1-\phi)^{1-r}
\end{equation}
Therefore, the posterior of $\phi$ also follows the Beta distribution:
\begin{equation}
    \phi|r \sim \text{Beta}(\alpha+r,\beta+1-r)
\end{equation}
Now let $n=\alpha+\beta$, plug these into \autoref{eqn:expected-variance-reduction-1}, and obtain:
\begin{align}
     \Delta V(\tau)  &= \text{Var}_{\text{prior}}(\phi_{\tau}) -\sum_{r'\in\{0,1\}}\text{Pr}(r=r')\text{Var}_{\text{posterior}}(\phi_{\tau}|r=r')\\
     &= \frac{\alpha\beta}{n^2(n+1)} - [\frac{\alpha}{n}\cdot \frac{(\alpha+1)\beta}{(n+1)^2(n+2)}+\frac{\beta}{n}\cdot \frac{\alpha(\beta+1)}{(n+1)^2(n+2)}]\\
     &=\frac{\alpha\beta}{n^2(n+1)} - \frac{\alpha\beta}{n(n+1)^2}\\
     &=\frac{\alpha\beta}{n^2(n+1)^2}
     \label{eqn:DeltaV}
\end{align}

\section{Derivation of Entropy of the Beta Distribution}
\label{appdix:entropy-beta-distribution}
Let $\Phi\sim \text{Beta}(\alpha,\beta)$ be a continous random variable with density:
\begin{equation}
    f(\phi)=\frac{1}{\text{B}(\alpha,\beta)}\phi^{\alpha-1}(1-\phi)^{\beta-1}
\end{equation}
where $\text{B}(\alpha,\beta)$ is the beta function. By definition of entropy, we have:
\begin{equation}
    H(\Phi) = -\mathbb{E}[\ln{f(\Phi)}]
\end{equation}
Then we can simplify it as:
\begin{equation}
    H(\Phi) = \ln{\text{B}(\alpha,\beta)} -(\alpha-1)\mathbb{E}[\ln{\Phi}]-(\beta-1)\mathbb{E}[\ln{(1-\Phi)}]
\end{equation}
The expectation $\mathbb{E}[\ln{\Phi}]$ and $\mathbb{E}[\ln{(1-\Phi)}]$ admit closed-form expressions in terms of the digamma function $\psi(\cdot)$:
\begin{equation}
    \mathbb{E}[\ln{\Phi}] = \psi(\alpha)-\psi(\alpha+\beta), \mathbb{E}[\ln{(1-\Phi)}]=\psi(\beta)-\psi(\alpha+\beta)
\end{equation}
Substituting these identities and simplifying gives the closed-form entropy:
\begin{equation}
\begin{aligned}
     H(\Phi)&=\ln{\text{B}(\alpha,\beta)} - (\alpha-1)(\psi(\alpha)-\psi(\alpha+\beta))-(\beta-1)(\psi(\beta)-\psi(\alpha+\beta)) \\
     &=\ln{\text{B}(\alpha,\beta)} - (\alpha-1)\psi(\alpha)-(\beta-1)\psi(\beta)+(\alpha+\beta-2)\psi(\alpha+\beta)
\end{aligned}
\end{equation}
\clearpage

\section{Proof of Asymptotic Scaling of Mutual Information}
\label{appdix:asymptotic-scaling-mutual-information}
For a sufficiently large $n_{\tau}=\alpha_{\tau}+\beta_{\tau}$, the Beta distribution would converge to a Gaussian distribution. The entropy of a Gaussian variable is given by:
\begin{equation}
H(\Phi_{\tau}) \approx \frac{1}{2} \ln(2\pi e \cdot \text{Var}(\Phi_{\tau})) 
\end{equation}
Then the Mutual Information can be approximated as the expected log-ratio of variance:
\begin{equation} 
I(R; \Phi_{\tau}) \approx \frac{1}{2} \mathbb{E}_R \left[ \ln \left( \frac{\text{Var}(\Phi{\tau})}{\text{Var}(\Phi_{\tau}|R)} \right) \right] \label{eqn:mi-log-var} \end{equation}

During RL training, when $n_\tau$ is large, the posterior variance $\text{Var}(\Phi_{\tau}|R)$ concentrates tightly around its expectation. Denoting the expected variance reduction by:
\begin{equation}
    \Delta V(\tau) = \mathbb{E}_R\left[\text{Var}(\Phi_{\tau})-\text{Var}(\Phi_{\tau}|R) \right]
\end{equation}

For large $n_{\tau}$, the ratio $\frac{\Delta V(\tau)}{\text{Var}(\Phi_{\tau})}$ is small. Then apply the Taylor expansion $\ln\left(1/(1-x)\right)\approx x$ for $x\approx 0$:
\begin{equation}
    \begin{aligned}
        I(R; \Phi_{\tau}) &\approx \frac{1}{2} \mathbb{E}_R \left[ \ln \left( \frac{\text{Var}(\Phi{\tau})}{\text{Var}(\Phi_{\tau}|R)} \right) \right] \\
        &\approx\frac{1}{2}\left[ \ln \left( \frac{\text{Var}(\Phi{\tau})}{\text{Var}(\Phi_{\tau})-\Delta V(\tau)} \right) \right] \\
        &= \frac{1}{2} \left[ \ln \left( \frac{1}{1-\frac{\Delta V(\tau)}{\text{Var}(\Phi_{\tau})}} \right) \right] \\
        &\approx \frac{1}{2} \left[ \left( \frac{\Delta V(\tau)}{\text{Var}(\Phi_{\tau})} \right) \right] 
    \end{aligned}
\end{equation}
Now substituting the exact expressions from \autoref{eqn:beta-variance} and \autoref{eqn:DeltaV} yields:
\begin{equation}
    \begin{aligned}
        I(R; \Phi_{\tau}) &\approx \frac{1}{2} \cdot \frac{\alpha_\tau\beta_\tau}{n_{\tau}^2(n_{\tau}+1)^2}\cdot \frac{n_{\tau}^2(n_{\tau}+1)}{\alpha_\tau\beta_\tau}\\
        &=\frac{1}{2(n_{\tau}+1)} \sim \mathcal{O}(\frac{1}{n_{\tau}})
    \end{aligned}
\end{equation}

\section{Implementation Details}
\label{appdix:implement-details}
\subsection{Datasets Details}
\noindent\textbf{CountDown.}  We adopt the Countdown Number Game, which requires combining given numbers using basic arithmetic operations to reach a target value. In detail, we randomly selected 2,000 samples from the CountDown-34 dataset\footnote{\url{https://huggingface.co/datasets/BlackBeenie/simple-countdown}} for training, and evaluated performances on a 512-problem held-out test set. In CountDown-34, each problem provides either 3 or 4 source numbers, and the distribution is balanced. We set the reward function as the follows:
\begin{equation}
r =
\begin{cases}
1,   & \text{if response is correct}, \\
0.1, & \text{if response is incorrect but with correct formatting}, \\
0,   & \text{otherwise}.
\end{cases}
\end{equation}

\noindent\textbf{DeepScaler.} We train LLMs on the training set of the DeepScaleR dataset\footnote{\url{https://huggingface.co/datasets/agentica-org/DeepScaleR-Preview-Dataset}} \citep{deepscaler2025}, which consists of approximately 40,000 unique mathematics problem-answer pairs. For evaluation,  we conduct it on a suite of math benchmarks: AIME24, AMC23, MATH500 \citep{lightman2023lets}, Minerva Math \citep{lewkowycz2022solvingquantitativereasoningproblems}, and OlympiadBench \citep{he2024olympiadbenchchallengingbenchmarkpromoting}, using the datasets provided by DeepScaler \citep{deepscaler2025}. We report the average \texttt{pass@1} over 16 samples for each problem. Following the default setting in \citet{Sheng_2025},we use a binary reward function that assigns a reward of 1 for a correct answer and 0 otherwise.

\noindent\textbf{WebInstruct-verified.} We train LLMs on the training set of WebInstruct-verified \citep{ma2025generalreasoner}, which is a diverse and high-quality dataset to enhance robust reasoning capabilities across a broad range of domains, including physics, chemistry, and more. For evaluation, we apply the LM-Evaluation-Harness \citep{eval-harness}, a unified framework to test generative language models on different evaluation tasks. For RL training, we adopt the rule-based reward function as used in the above mathematics task.

\subsection{Training Details}
We adopt the widely used GRPO \citep{shao2024deepseekmathpushinglimitsmathematical}, implemented in the VeRL framework \citep{Sheng_2025}, as our default algorithm. It is worth noting that our method \textsc{InSight} can be integrated into any other RLVR algorithms.

We set training steps as $100$ during RL training. At each training step, we set $k = 8$ responses per prompt for estimating advantages, using temperature $1.0$ and top p = 1.0. Evaluation is based on \texttt{pass@1}, computed from $16$ independent generations per prompt with temperature $0.6$ and top p = 0.95, keeping the same setting as in \citet{deepscaler2025}. Training batch sizes are set to 256, while the mini-batch sizes are 128 and 64 for Mathematics\&General-reasoning and Countdown, respectively. Optimization is performed by AdamW \citep{loshchilov2019decoupledweightdecayregularization}, with learning rate $1e-6$, weight decay $0.1$, beta $(0.9, 0.999)$. For online data selection methods, (including \textsc{MoPPS}, \textsc{InSight}, \textsc{Inverse-Evidence}, \textsc{Expected-Difficulty}), we set $\hat{M}$ is 16x of the selected batch size $M$, and $\lambda=1.0$. For our \textsc{InSight}, we set $\eta=3.0,\mu=3.0$. All experiments are conducted on 8x NVIDIA H-series GPUs.

\section{Performance and Runtime Comparisons Between \textsc{InSight} and Dynamic Sampling}
\label{appdix:performance-runtime-comparison}
In this section, we compare the performance and runtime (i.e., total training hours) between \textsc{InSight} and Synamic Sampling (DS), where the training setting is the same as described above. \textsc{InSight} precomputes the WMI scores for each datapoint, which is significantly faster than DS, since DS requires the LLM to over-sample the data and keep only effective datapoints after exact generation and evaluations. 

\begin{table*}[!ht]
\centering
\caption{Evaluation results across mathematics benchmarks. Accuracy is computed as the average \texttt{pass@1} over 16 independent generations per problem.} 
\label{tab:performance-runtime-comparison}
\resizebox{\linewidth}{!}{%
\begin{tabular}{llllllll}
\toprule
\textbf{\textsc{Models}}                      & \textbf{\textsc{Methods}}           & \textbf{AIME24}              & \textbf{\textsc{AMC23}}               & \textbf{MATH500}             & \textbf{\textsc{Minerva.}}            & \textbf{\textsc{Olympiad.}}           & \textbf{\textsc{Training Hours}}                \\ \midrule
                
                                     & {\color[HTML]{9B9B9B} DS}  &{\color[HTML]{9B9B9B} 50.62} & {\color[HTML]{9B9B9B} 80.79} & {\color[HTML]{9B9B9B} 90.82} & {\color[HTML]{9B9B9B} 38.39} & {\color[HTML]{9B9B9B} 53.38} & {\color[HTML]{9B9B9B} $\sim$ 30.5 Hours } \\
                                 \textsc{R1-Distill-Qwen-7B}       & \textsc{Random}                      & 46.25                        & 77.90                        & {90.50 }                       & 37.98                        & 51.80                        &  $\sim$ 12.5 Hours                     \\

 &  \textcolor{Tomato}{\textsc{InSight} (Ours)}                               &47.71              & {78.40}                & 90.45                       & \textbf{39.06}                         & {52.50}                 &   $\sim$ 12.5 Hours                      \\ \bottomrule
\end{tabular}
}
\end{table*}

\noindent\textbf{Analysis.} \autoref{tab:performance-runtime-comparison} shows the comparisons on R1-Distill-Qwen-7B, where we set the \texttt{trainer.total\_training\_steps=100} in VeRL \citep{Sheng_2025} training for fair comparison. As expected, DS incurs substantially higher computational cost, requiring more than \textbf{2x} the total training hours of \textsc{InSight}. This overhead arises from DS's reliance on oversampling, followed by full generation and reward evaluation to filter effective datapoints.

Despite this additional cost, DS achieves only marginal performance gains over \textsc{InSight} and, in some benchmarks, performs comparably or even worse. In contrast, \textsc{InSight} consistently improves over \textsc{Random} while matching or exceeding DS on most metrics, without introducing any extra training-time overhead. These results highlight that information-aware selection based on Bayesian uncertainty can recover much of the benefit of oversampling-based methods, while being significantly more computationally efficient.

Overall, this comparison underscores the practical advantage of \textsc{InSight}: it achieves strong performance gains comparable to expensive dynamic sampling strategies, yet preserves the runtime efficiency of standard RL training pipelines.

\section{Ablation Analysis of the Effect of Larger Candidate Batch Size $\hat{M}$}
\label{appdix:hatM-analysis}
In this section, we further study the effect of the larger candidate batch size $\hat{M}$ in Algorithm~\autoref{alg:wmi-algorithm}. We conduct experiments on $\hat{M}=$ 8x, 12x, and 16x (our original setting) to compare the performances.

\begin{table}[]
\centering
\caption{Evaluation comparisons of different $\hat{M}$ across different models, on mathematics benchmarks. \textbf{Bold} indicates the best results.} 
\label{tab:hatM-analysis}
\resizebox{0.9\linewidth}{!}{%
\begin{tabular}{llcccccc}
\toprule
\textbf{\textsc{Models}}    &\textbf{\textsc{Methods}}                  & $\hat{M}$                        & \textbf{AIME24}                   & \textbf{\textsc{AMC23}}           & \textbf{MATH500} & \textbf{\textsc{Minerva.}} & \textbf{\textsc{Olympiad.}}                \\ \midrule
       & \textsc{MoPPS} &   8x & 10.20 & 44.42  & 71.46 & 21.64&34.07  \\
     & \textsc{InSight} & 8x & 10.20 & \textbf{44.50} & \textbf{72.53} & 21.69 & \textbf{35.07} \\
   \textsc{Qwen3-0.6B} & \textsc{InSight} & 12x & 10.83 & 43.30& 70.85 & \textbf{22.20} & 34.33 \\
     & \textsc{InSight} & 16x & \textbf{11.90}	&44.30	&71.75	&21.94	&34.80 \\ \midrule
    & \textsc{MoPPS} & 8x & 52.29 &  77.71 & 91.08 & 41.40  & 56.70\\
      & \textsc{InSight} & 8x & 52.29 & 78.54 & 91.09 & 41.73 & 56.91  \\
    \textsc{Qwen3-4B}   & \textsc{InSight} & 12x & 51.25 & 78.31 & \textbf{91.27} & 41.43 & 56.92 \\
        & \textsc{InSight} & 16x & \textbf{53.75}	&\textbf{79.00}	&91.22	&\textbf{41.80}	&\textbf{57.00}\\ \midrule
    & \textsc{InSight} & 8x & 46.45 & 77.10  & 90.33 &  38.37 & 51.98 \\
    \textsc{R1-Distill-7B}   & \textsc{InSight} & 12x & 46.67 & 78.01 & \textbf{90.48} & 38.46 & \textbf{52.51} \\
      & \textsc{InSight} & 16x & \textbf{47.71}	&\textbf{78.40}	&90.45	&\textbf{39.06}	&52.50\\ \bottomrule
\end{tabular}
}
\end{table}

\noindent\textbf{Analysis.} \autoref{tab:hatM-analysis} examines the effect of the larger candidate batch size $\hat{M}$ on mathematical reasoning benchmarks across different model scales. We observe a non-monotonic trend for smaller models and a more consistent preference for larger $\hat{M}$ as model size increases.

For \textsc{Qwen3-0.6B}, intermediate candidate sizes (8x or 12x) achieve the best performance on several benchmarks, occasionally outperforming 16x. This behavior suggests that, for smaller models, excessively large candidate pools may introduce additional variance or overly challenging samples that exceed the model’s effective learning capacity. In this regime, a moderately sized candidate set strikes a better balance between exploration and learnability, allowing \textsc{InSight} to select informative yet tractable datapoints.

In contrast, as model capacity increases (e.g., \textsc{Qwen3-4B} and \textsc{R1-Distill-7B}), performance improves more consistently with larger $\hat{M}$, and 16x becomes the strongest setting across most benchmarks. Larger models are better able to exploit increased candidate diversity and accurately estimate epistemic uncertainty, enabling more reliable ranking of high-information datapoints from a broader pool.

Overall, these results indicate that the optimal $\hat{M}$ depends on model capacity: smaller models benefit from moderate candidate sizes that avoid over-exploration, while larger models can effectively leverage larger pools to improve selection quality. This adaptive behavior further highlights the robustness of \textsc{InSight} across model scales and practical training settings.

\end{document}